\documentclass[journal]{article}
\usepackage{cite}
\usepackage{amsmath,amssymb,amsfonts}
\usepackage{algorithm2e}
\usepackage{graphicx}
\usepackage{textcomp}
\usepackage{array}
\usepackage{graphicx}
\usepackage{caption}
\usepackage{subcaption}
\usepackage{enumerate}
\usepackage{booktabs}
\usepackage{multirow}
\usepackage{threeparttable}
\usepackage{verbatim}
\usepackage{placeins}
\newcommand{\myparagraph}[1]{\vspace{0.1em}\noindent\textbf{#1}}
\usepackage[colorlinks,hyperindex,breaklinks]{hyperref}
\usepackage[noend]{algpseudocode}
\usepackage[font=small,labelfont=bf,justification=justified,format=plain]{caption} 
\newcommand{\ie}{\textit{i}.\textit{e}.}
\newcommand{\eg}{\textit{e}.\textit{g}.}
\newcommand{\etal}{\textit{et al}.}
\begin{document}
\title{FedMix: Mixed Supervised Federated Learning for Medical Image Segmentation}
\author{Jeffry Wicaksana, Zengqiang Yan, Dong Zhang, Xijie Huang, Huimin Wu,
		\\Xin Yang, and Kwang-Ting Cheng
\thanks{J. Wicaksana, D. Zhang, X. Huang, H. Wu and K. -T. Cheng are with the Department of Electronic and Computer Engineering, The Hong Kong University of Science and Technology, Kowloon, Hong Kong. E-mail: \{jwicaksana, dongz, xhuangbs, hwubl, timcheng\}@ust.hk}
\thanks{Z. Yan and X. Yang are with the School of Electronic Information and Communications, Huazhong University of Science and Technology, Wuhan, China. E-mail: \{z\_yan ,xinyang2014\}@hust.edu.cn}}
\maketitle
\begin{abstract} 
The purpose of federated learning is to enable multiple clients to jointly train a machine learning model without sharing data. However, the existing methods for training an image segmentation model have been based on an unrealistic assumption that the training set for each local client is annotated in a similar fashion and thus follows the same image supervision level. To relax this assumption, in this work, we propose a label-agnostic unified federated learning framework, named FedMix, for medical image segmentation based on mixed image labels. 
In FedMix, each client updates the federated model by integrating and effectively making use of all available labeled data ranging from strong pixel-level labels, weak bounding box labels, to weakest image-level class labels. Based on these local models, we further propose an adaptive weight assignment procedure across local clients, where each client learns an aggregation weight during the global model update. 
Compared to the existing methods, FedMix not only breaks through the constraint of a single level of image supervision, but also can dynamically adjust the aggregation weight of each local client, achieving rich yet discriminative feature representations. To evaluate its effectiveness, experiments have been carried out on two challenging medical image segmentation tasks, \ie, breast tumor segmentation and skin lesion segmentation. The results validate that our proposed FedMix outperforms the state-of-the-art methods by a large margin\footnote{The code is available at: https://github.com/Jwicaksana/FedMix}.
\end{abstract}

\section{Introduction} 
\label{sec:introduction}
Medical image segmentation is a 
representative task for image content analysis supporting computer aided diagnosis, which can not only recognize the lesion category, but also locate the specific areas~\cite{long2015fully}. In the past few years, this task has been extensively studied and applied in a wide range of underlying scenarios, \eg, lung nodule segmentation~\cite{jin2018ct}, skin lesion boundary detection~\cite{codella2018skin}, and COVID-19 lesion segmentation~\cite{li2021dual}. 

The optimization of deep learning models usually relies on a vast amount of training data~\cite{he2016deep}. For example, for a fully-supervised semantic segmentation model, the ideal scenario is that we can collect the pixel-level annotated images as much as possible from diverse sources. However, this scenario is almost infeasible due to the following two reasons: $\romannumeral1$) the strict sharing protocol of sensitive patient information between medical institutions and $\romannumeral2$) the exceedingly high pixel-level annotation cost. As the expert knowledge usually required for annotating medical images is much more demanding and difficult to obtain, various medical institutions have very limited strong pixel-level annotated images and most available images are unlabeled or weakly-annotated~\cite{li2021dual,anno_all,zhang2019}. Therefore, a realistic clinical mechanism which utilizes every available supervision for cross-institutional collaboration without data sharing is highly desirable. 
\begin{figure}[t!]
\centering
\includegraphics[width=1\columnwidth]{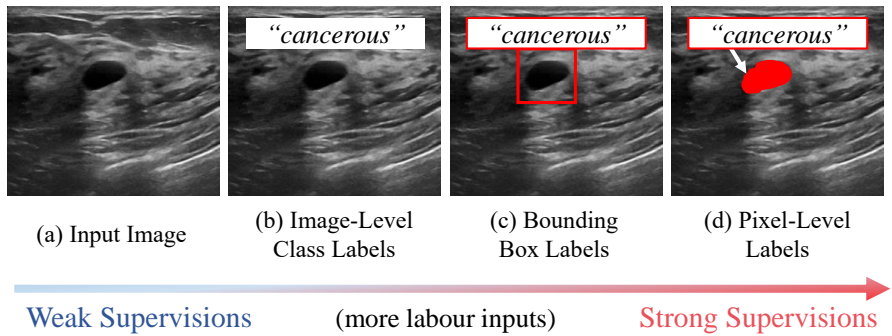}
\caption{Examples of different levels of medical image labels, where the image-level class labels in (b) contain only the lesion category. The bounding box labels in (c) contain not only the lesion category, but also a coarse location. The pixel-level labels in (d) contain both the lesion category and location information of each pixel, which is a strong image supervision. Although strong image supervisions are more informative, they are very expensive to obtain. The utilization of some easily obtained image supervisions is beneficial in practice.}
\label{overview_lay}
\end{figure}
\begin{figure*}[t]
\centering
\includegraphics[width=.85\textwidth]{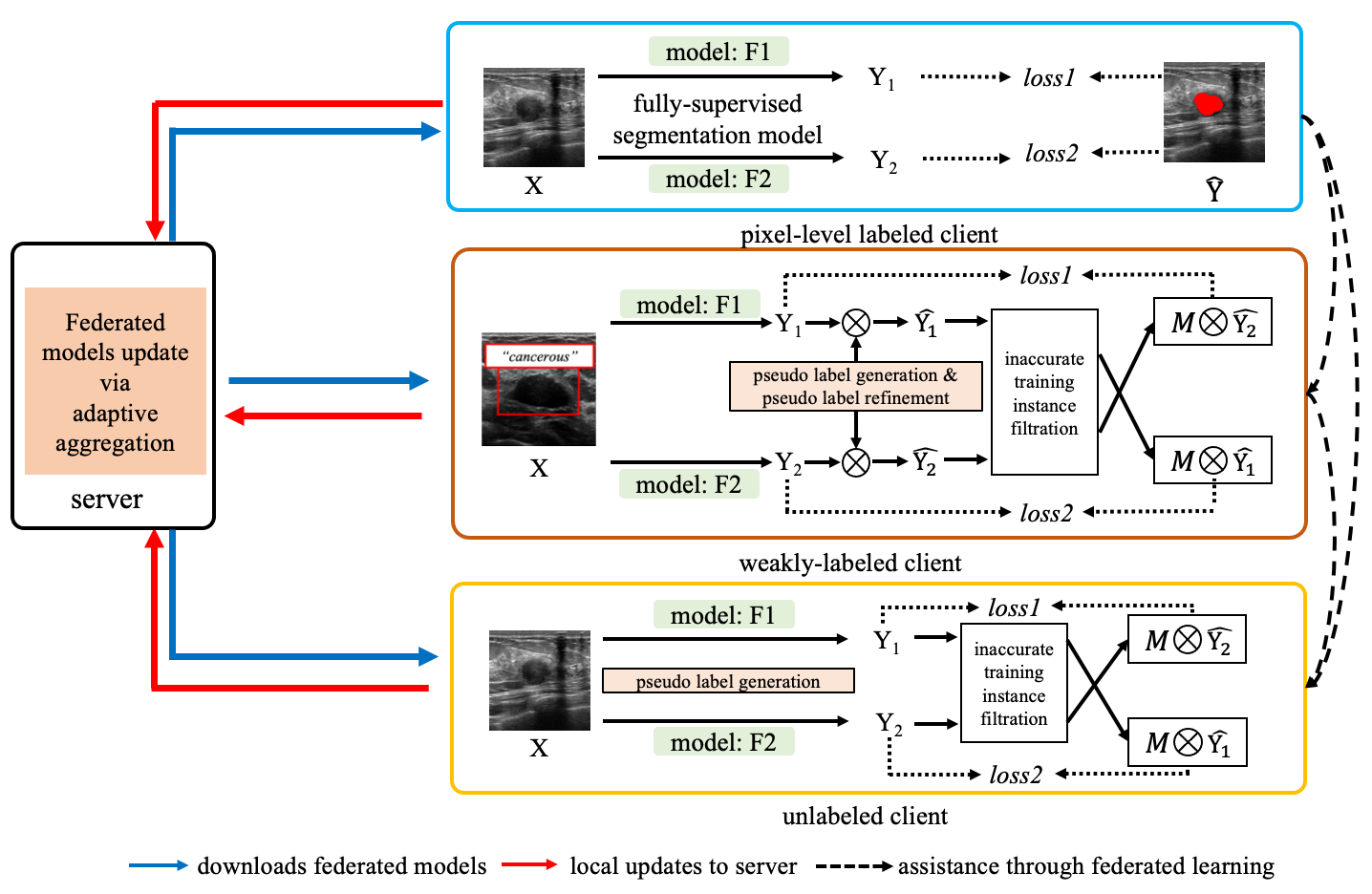}
\caption{An illustration of our proposed Mixed Supervised Federated Learning (FedMix) framework. The local client update utilizes every available supervision for training. Based on which, an adaptive weight aggregation procedure is used for the global federated model update. Compared to the existing methods, FedMix not only breaks through the constraint of a single level of image supervision, but also can dynamically adjust the aggregation weight of each local client, achieving a rich yet discriminative feature representation.}
\label{overview}
\end{figure*}

Thanks to the timely emergence of Federated Learning (FL), which aims to enable multiple clients to jointly train a machine learning model without sharing data, the problem of data privacy being breached can be alleviated~\cite{fl}. FL has gained significant attention in the medical imaging community~\cite{sheller, fl_pp_brain}, due to the obvious reason that medical images often contain some personal information. During the training process of a standard FL model, each local client first downloads the federated model from a server and updates the model locally. Then, the locally-trained model parameters of each client are sent back to the server. Finally, all clients' model parameters are aggregated to update the global federated model. Most of the existing FL frameworks~\cite{fmri_abide, dou_npj} require that the data used for training by each local client needs to follow the same level of labels, \eg, {pixel-level labels} (as shown in Fig.~\ref{overview_lay} (d)) for an image semantic segmentation model, which limits the model learning ability. Although, some semi-supervised federated learning methods~\cite{ssl_seg_fl, fedperl} attempt to utilize the unlabeled data in addition to pixel-level labeled images in training, they do not make any use of the weakly-labeled images (\eg, {image-level class labels} in Fig.~\ref{overview_lay} (b) and {bounding box labels} in Fig.~\ref{overview_lay} (c)), which are invaluable.   

Clients participating in FL may have different labeling budgets. Therefore, there may be a wide range of inter-client variations in label availability. Weak labels are easier to acquire and thus more broadly available compared to pixel-level ones. In practice, there is a wide range of weak labels with varying strengths and acquisition costs. While an image-level label indicating whether a breast ultrasound image is cancerous or not is easier to acquire compared to a bounding box label pointing out the specific location of the cancerous region, it is also less informative. Therefore, effectively utilizing the information from these weakly-labeled data with varying levels of label strengths as well as unlabeled data, especially for clients without pixel-level labeled data would be highly beneficial for improving the federated model's robustness while preventing training instability. 

In this work, as illustrated in Fig.~\ref{overview}, we propose a label-agnostic Mixed Supervised Federated Learning (FedMix) framework, which is a unified FL model making use of data labeled in any form for medical image segmentation. Specifically, in the absence of {pixel-level labels}, FedMix first effectively utilizes unlabeled images as well as useful information contained in the weakly-labeled images (\ie, {image-level class labels} and {bounding box labels}) for producing and selecting high-quality pseudo labels. Through an iterative process, the accuracy of selected pseudo labels which are then used for local training on the client sides improves, leading to better model performance. 
To further improve the model robustness, FedMix takes into account the variability of local clients' available labels through an adaptive aggregation procedure for updating the global federated model. Compared to the existing methods, FedMix not only breaks through the constraint of a single type of labels, but also can dynamically assign an optimized aggregation weight to each local client. Experimental results on two challenging segmentation tasks demonstrate the superior performance of FedMix on learning from mixed supervisions, which is valuable in the clinical setting. Our contributions are summarized as follows:
\begin{itemize}
	\item The mixed supervised FL framework targeting multi-source medical image segmentation through an iterative pseudo label generator followed by a label refinement operation, based on the information derived from weakly-labeled data, to target high-quality pseudo labels for training.
	\item An adaptive weight assignment across clients, where each client can learn an aggregation weight. Adaptive weight assignment is essential to handle inter-client variations in supervision availability. 
	\item Extensive experiments on the challenging breast tumor segmentation and skin lesion segmentation. FedMix outperforms the state-of-the-art methods by a large margin.
\end{itemize}

The rest of this paper is organized as follows: Existing and related work are summarized and discussed in Section~\ref{lit}. The details of FedMix are introduced in Section~\ref{framework}. In Section \ref{evaluation}, we present thorough evaluation of FedMix compared with the existing methods. We provide ablation studies as well as analysis in Section~\ref{discussion}, and conclude the paper in Section~\ref{conclusion}.
\section{Related Work}\label{lit}
\subsection{Federated Learning}
Federated learning (FL) is a distributed learning framework, which is designed to allow different clients, institutions, and edge devices to jointly train a machine learning model without sharing the raw data~\cite{fl}, which plays a big role in protecting data privacy. In the past years, FL has drawn great attention from the medical image communities~\cite{fl_natural, dou_npj} and has been validated for multi-site functional magnetic resonance imaging classification~\cite{fmri_abide}, health tracking through wearables~\cite{fl_wearable}, COVID-19 screening and lesion detection~\cite{fl_covid}, and brain tumor segmentation~\cite{fl_pp_brain, sheller}. In clinical practice, different clients may have great variations in data quality, quantity, and supervision availability. 
Improper use of these data may lead to significant performance degradation among different clients. To reduce the inter-client variations, FL has been combined with domain adaptation~\cite{eegdomain, feddg, VAFL}, contrastive learning~\cite{moon} and knowledge distillation~\cite{fedmd} to learn a more generalizable federated model. However, existing works do not consider the variation in supervision availability (\ie, different clients have different levels of image labels), which is often observed in clinical practice. In our work, we use all the available image label information including image-level class labels, bounding box labels, and pixel-level labels to train a medical image segmentation model and propose a mixed supervised FL framework.

\subsection{Semi-supervised Federated Learning}
In a standard federated learning setting, not every local client has access to pixel-level supervision for image segmentation to facilitate model learning with weakly-labeled and unlabeled training data. To this end, some semi-supervised federated learning approaches require clients to share supplementary information, \eg, client-specific disease relationship~\cite{fed_irm}, extracted features from raw data~\cite{fed_volum}, metadata of the training data~\cite{fed_con}, and ensemble predictions from different clients' locally-updated models besides their parameters~\cite{fedperl}. Additional information sharing beyond the locally-updated model parameters may leak privacy-sensitive information~\cite{leakage} about clients' data. Yang \etal~\cite{ssl_seg_fl} proposed to avoid additional information sharing by first training a fully-supervised federated learning model only on clients with available pixel-level supervision for several training rounds and then using the model to generate pseudo labels for local clients based on the unlabeled data. Those confident pseudo labels are used to supervise the local model updates on unlabeled clients for subsequent rounds. In this work, we design a unified federated learning framework that utilizes various weakly supervised data in addition to fully-supervised and unlabeled data for training while limiting the information sharing between clients to only locally-updated model parameters for privacy preservation.

\subsection{Medical Image Segmentation}
The deep learning-based image recognition technology has been used for various medical image segmentation tasks, \eg, optic disc segmentation~\cite{oc}, lung nodules segmentation~\cite{jin2018ct}, skin lesion boundary detection~\cite{codella2018skin}, and COVID-19 lesion segmentation~\cite{li2021dual}. However, training a fully-supervised deep model for image semantic segmentation often requires access to a mass of pixel-level supervisions, which are expensive to acquire~\cite{anno_all}. In particular, the problem of the expensive pixel-level supervision is much more obstructive for medical image segmentation~\cite{xiaomeng_rot}. To this end, efforts have been made to explore the use of some easily obtained image supervisions (\eg, scribbles~\cite{scribble_can}, image-level classes~\cite{ahn2018learning}, bounding boxes~\cite{dai2015boxsup}, points~\cite{bearman2016s}, and even unlabeled image\cite{cps}) to train a pixel-level image segmentation model. However, most of the existing works are based on only one or two types of image supervisions, which greatly limits the model learning efficiency. In most cases, access to some pixel-level annotated data is required to facilitate model training, which may not always be available for each participating client. In our work, we carefully use image-level class labels, bounding box labels, and pixel-level labels to train local clients and propose an adaptive weight assignment procedure across clients for medical image segmentation.
\section{Our Approach}\label{framework} 
In this section, we first introduce the notation and experimental setting of the proposed unified federated learning framework, \ie, Fedmix, in Section~\ref{sec3:1}. Then, we provide a framework overview in Section~\ref{sec3:2}. Finally, we present implementation details including pseudo label generation, selection, and federated model update of the proposed FedMix in Section~\ref{sec3:3} and Section~\ref{fl_aggr}.

\subsection{Preliminaries}\label{sec3:1}
\subsubsection{Experimental Settings}
To emulate the real scenario setting, we focus on deep learning from multi-source datasets, where each client's data is collected from different medical sources. We focus on exploring variations in cross-client supervisions and thus limit each client to a single level of labels. 

\subsubsection{Training Notations}
In this paper, we denote $\overline D = [D_{1}, ..., D_{N}]$ as the collection of $N$ clients' training data. Given client $i$, $D_{i}^{L} = [X, Y_{gt}]$, $D_{i}^{U}=[X]$, $D_{i}^{img}=[X, Y_{img}]$, and $D_{i}^{bbox}=[X, Y_{bbox}]$ represent the training data that is pixel-level labeled, unlabeled, image-level class labeled, and bounding box-level labeled, respectively. $X$ and $Y$ represent the sets of the training images and the available labels.

To integrate various levels of image labels, in our work, we modify the bounding box labels and image-level class labels to pixel-level labels. Specifically, the bounding box point representation is converted into pixel-level label where the foreground class falls inside the bounding box and the background class falls outside the bounding box. For image-level class labels, we constrain the pixel-level label to the corresponding image class. Consequently, $Y_{gt}$, $Y_{img}$, and $Y_{bbox}$ has the same dimension, \eg, $Y \in \mathbb{R}^{(C+1)\times H \times W}$, $C$ indicates the total number of foreground classes while $W$ and $H$ indicates the weight and height of the respective image data.

\subsection{Overview}\label{sec3:2}
As illustrated in Fig. \ref{overview}, to fully utilize \emph{every level of labels} at various clients, the pseudo-code of FedMix is presented in Algorithm~\ref{alg} and FedMix has two main components: 
\begin{enumerate}
	\item \textbf{Pseudo Label Generation and Selection.} In the mixed supervised setting, clients without access to pixel-level label rely on the pseudo labels for training. To improve the pseudo labels' accuracy, we design a unified refinement process using \emph{every level of labels} and dynamically select high-quality pseudo labels for training.
	\item \textbf{Adaptive Aggregation for Federated Model Update.} FedMix uses an adaptive aggregation operation where the weight of each client is determined by not only its data quantity but also the quality of its pseudo labels. Our aim is to learn a federated model for tumor segmentation, the local model updates without access to pixel-level labels have to be integrated with care. In this way, the reliable clients will be assigned higher aggregation weights, leading to a better federated model. 
\end{enumerate} 

\RestyleAlgo{ruled}
	\SetKwComment{Comment}{/* }{ */}

\begin{algorithm}[t]
	\caption{Pseudocode of FedMix}\label{alg}
	\SetKwInOut{Input}{input}
	\SetKwInOut{Output}{output}
	\SetKwInOut{Parameter}{parameter}
	\Input{$\overline D$}
	\Parameter{$\beta$, $\lambda$: hyperparameters for adaptive aggregation\\ $T$: maximum training rounds\\ $\epsilon$: threshold for dynamic sample selection}
	\Output{$\theta_{\xi1}$: parameters of $F_{1}$\\ $\theta_{\xi2}$: parameters of $F_{2}$}
	
	$\theta_{\xi1}$, $\theta_{\xi2}$ $\leftarrow$ initialize()\\
	\For{$t=1:T$}{
		${\cal L}^{t} = \{\}$, 
		${\theta}^{t}_{\xi1} = \{\}$, 
		${\theta}^{t}_{\xi2} = \{\}$\\
		\For{$i=1:N$}{
			$f^i_{1}, f^i_2$ $\leftarrow$ download($\theta_{\xi1}$, $\theta_{\xi2}$) \\
			$X, Y \leftarrow D_i$ \\
			$Y_{1}, Y_{2} \leftarrow F^i_{1}(X), F^i_{2}(X)$ \\
			$M_{i} \leftarrow \textbf{sample}(Y_{1}, Y_{2}, \epsilon)$ \\
			$\hat Y_{1}, \hat Y_{2} \leftarrow \textbf{refine}(Y_{1}, Y_{2}, Y)$ \\
			$d_{i} \leftarrow M_{i} * D_{i}$ \\
			$\Delta {\theta}^{t}_{i1}$, $\Delta {\theta}^{t}_{i2}$, ${\cal L}^{t}_i$ $\leftarrow$ update($F^i_{1}, F^i_{2}; d_i$)\\
			${\theta}^{t}_{\xi1}$.add($\Delta {\theta}^{t}_{i1}$), 
			${\theta}^{t}_{\xi2}$.add($\Delta {\theta}^{t}_{i2}$), 
			${\cal L}^{t}$.add(${\cal L}^{t}_i$)\\
		}
		${\theta}_{\xi1}$, ${\theta}_{\xi2}$  $\leftarrow$ \textbf{aggregate}(${\theta}^{t}_{\xi1}$, ${\theta}^{t}_{\xi2}$, ${\cal L}^{t}$; $\beta$, $\lambda$)\\
	}
	return $\theta_{\xi1}$ and $\theta_{\xi2}$
\end{algorithm}

\subsection{Pseudo Label Generation and Selection}\label{sec3:3}
\subsubsection{Pseudo Label Generation}
Based on the cross-pseudo supervisions~\cite{cps}, we train two differently initialized models, $F_{1}(.)$ and $F_{2}(.)$ to co-supervise each other with pseudo labels when no pixel-level label is available. 
The training image $X$ is fed to the two models $F_{1}$ and $F_{2}$ to generate pseudo labels $Y_{1}$ and $Y_{2}$, respectively. The pseudo labels are then refined, denoted as ${\hat Y}_{1}$ and ${\hat Y}_{2}$, and used for training the model of each local client. Details of the corresponding refinement strategies for each type of label are introduced as follows: 

\begin{enumerate}

	\item \textbf{Pixel-level labels}: Under this kind of supervision, we do refine the pseudo labels, which can be expressed as ${\hat Y}_1={\hat Y}_2=Y_{gt}$. 
	\item \textbf{Bounding box labels}: 
	Each of the predictions $Y_1=F_1(X_1)$ and $Y_2=F_2(X_2)$ is refined according to the corresponding bounding box label, \ie, ${\hat Y}_1=Y_1*Y_{bbox}$ and ${\hat Y}_2=Y_2*Y_{bbox}$.  
	\item \textbf{Image-level class labels}: We do not apply pseudo label refinement, which can be formulated as ${\hat Y}_1=Y_1$, and ${\hat Y}_2=Y_2$.
	\item \textbf{No labels} (\ie, without supervisions): We do not refine the pseudo labels, which is formulated as ${\hat Y}_1=Y_1$, and ${\hat Y}_2=Y_2$.
\end{enumerate}
A specific client $i$ is trained by minimizing:
\begin{equation}\label{overall}
	{\cal L}_{i} = {\cal L}_{dice}(Y_{1}, \hat Y_{2})+{\cal L}_{dice}(Y_{2}, \hat Y_{1}),
\end{equation}
where ${\cal L}_{dice}$ is the Dice loss function.

\subsubsection{Dynamic Sample Selection}
Despite the effectiveness of the above pseudo label generation and refinement processes, the pseudo labels may be incorrect. 
Therefore, we propose a dynamic sample selection approach to select high-quality data and pseudo labels. 
Specifically, given client $i$ and its training data $D_i$, we generate a mask $M_{i} = \{m_{1}, ... , m_{|D_{i}|} | m_{i} \in [0,1]\}$ to select reliable training samples according to Eq.~\ref{sample_sel}. We measure the consistency between pseudo labels before refinement, \ie, $Y_{1}$ and $Y_{2}$. Higher prediction consistency between $Y_{1}$ and $Y_{2}$ indicates a higher likelihood that the pseudo labels are closer to ground truth. The above process is expressed as:
\begin{equation}\label{sample_sel}
	m_{i} = \begin{cases}
		1 & \text{if } dice(Y_{1}, Y_{2}) >= \epsilon\\
		0 & \text{o.w.},
	\end{cases}
\end{equation}
where $\epsilon \in [0,1]$ is a threshold which is inversely proportional to the amount of selected training samples. 
For pixel-level labels, $m_i=1$ for all training samples as $\hat Y_{1} = \hat Y_{2} = Y_{gt}$. As training progresses, the models are more capable of generating more accurate pseudo labels. Consequently, $\sum_{i_=1}^{i=|M_{i}|} m_{i}$ progressively increases to $|D_{i}|$, allowing the model to learn from a growing set of training data. More discussions of dynamic sample selection are provided in Section~\ref{consistencies_reliable}. 

\subsection{Federated Model Update}\label{fl_aggr}
At each training round, every local client $i$ first receives the federated model's parameters $\theta^{t}_{\xi}$ from the server at time or iteration $t$. Then, every client updates the model locally with its training data $D_{i}$. Finally, the gradient update from each local client $\Delta \theta ^{t+1}_{i}$ will be sent to the server to update the federated model's parameters according to Eq.~\ref{fed_avg_eq}.
\begin{equation}\label{fed_avg_eq}
	\theta^{t+1}_{\xi} \leftarrow \theta^{t}_{\xi} + \sum_{i=1}^N w_{i} \Delta \theta ^{t+1}_{i}.
\end{equation}
In FedAvg \cite{fl}, the aggregation weight of each client, $w_{i}$, is defined as $|D_{i}| / \sum_{i=1}^{i=|\overline D|}|D_{i}|$. In the mixed supervised setting, relying only on data quantity for weight assignment is sub-optimal. Thus, supervision availability of each client should also be taken into account during the aggregation process. To this end, we propose to utilize the client-specific training loss to infer the data quality. Each client's training loss not only provides a more objective measurement of its importance during FedMix optimization but also prevents the federated model from relying on the over-fitting clients. The proposed adaptive aggregation function is defined by
\begin{equation}
\begin{split}
	c_{i} \leftarrow
	\dfrac{|D_{i}|}{\sum_{i=1}^{i=|\overline D|}|D_{i}|} ,
	d_{i} \leftarrow
	\dfrac{{\Delta \cal L}_{i}^{\beta} }{\sum_{i=1}^{i=|\overline D|} {\Delta \cal L}_{i}^{\beta}} ,
\end{split}
\end{equation}
and
\begin{equation}\label{adjust_client_weight}
	w_{i} \leftarrow
	\dfrac{c_{i} + \lambda \cdot d_{i}}
	{\sum_{i=1}^{i=|\overline D|} c_{i} + \lambda \cdot d_{i}},
\end{equation}
where $\lambda$ and $\beta$ are hyper-parameters to tune, impacting the degree of reliance towards different clients. More discussions of adaptive aggregation can be found in Section~\ref{value_aa}.
\section{Experiments} \label{evaluation}
\subsection{Datasets and Evaluation Metrics}
\myparagraph{Dataset.} In our work, experiments are carried out on two challenging medical image segmentation tasks: 
\begin{itemize}
    \item \textbf{Breast tumor segmentation.} In this task, three public breast ultrasound datasets, namely BUS~\cite{breast_1}, BUSIS~\cite{breast_2}, and UDIAT~\cite{breast_3}, are used for evaluation and each of them is regarded as a separate client. More details of this dataset are introduced in Table~\ref{breast_info}.
    \item \textbf{Skin tumor segmentation.} HAM10K~\cite{ham10k} consists of four different sources. Each source acts as a client in FL. The statistics of HAM10K are presented in Table~\ref{skin_info}. 
\end{itemize}
Following the standard practice, the training data is randomly and patient-wisely split into 80\% for training and 20\% for testing. All the breast ultrasound and skin dermoscopy images are resized to 256$\times$256 pixels and then randomly flipped and cropped to 224$\times$224 pixels for training. 

{\renewcommand{\arraystretch}{1.0}
	\begin{table}[t]
		\centering
		\caption{Statistics of the breast ultrasound dataset}\label{breast_info}
		\begin{tabular}{c|c|cc}
			\hline
			\multirow{2}{*}{Site} & \multirow{2}{*}{\# Patients} & \multirow{2}{*}{\# Images} & \# Healthy    \\ \cline{4-4} 
			&                              &                            & \# Cancerous \\ \hline
			\multirow{2}{*}{BUS}    & \multirow{2}{*}{600}         & \multirow{2}{*}{780}       & 133                 \\ \cline{4-4} 
			&                              &                            & 647                 \\ \hline
			\multirow{2}{*}{BUSIS}  & \multirow{2}{*}{562}         & \multirow{2}{*}{562}       & 0                   \\ \cline{4-4} 
			&                              &                            & 562                 \\ \hline
			\multirow{2}{*}{UDIAT}  & \multirow{2}{*}{163}         & \multirow{2}{*}{163}       & 0                   \\ \cline{4-4} 
			&                              &                            & 163                 \\ \hline
		\end{tabular}
	\end{table}
}

{\renewcommand{\arraystretch}{1.0}
	\begin{table}[t]
		\centering
		\caption{Statistics of the HAM10K dataset}\label{skin_info}
		\begin{tabular}{c|c|c|c}
			\hline
			Site                   & Source    & \# Patients & \# Images \\ \hline
			Rosendahl              & rosendahl & 1552        & 2259      \\ \hline
			\multirow{3}{*}{Vidir} & modern    & 1695        & 3363      \\
			& old       & 278         & 439       \\
			& molemax   & 3954        & 3954      \\ \hline
		\end{tabular}
	\end{table}
}

\myparagraph{Evaluation metrics.} In this work, Dice coefficient (DC) is used for the evaluation of the two segmentation tasks. Considering the two-model architecture of FedMix, the predictions or outputs of $F_{1}$ are used for evaluation. 

\subsection{Implementation Details}
\myparagraph{Network architectures.} UNet~\cite{unet} combined with the group norm~\cite{groupnorm} is selected as the baseline segmentation model. 

\myparagraph{Supervision types.} The following types of labels are included in our experiments: 1) pixel-level labels (denoted as $L$), 2) bounding box labels (denoted as $B$), 3) image-level class labels (denoted as $I$), and 4) unlabeled (denoted as $U$), \eg, training with only the raw images. 

\myparagraph{Comparison methods.}  The following four prevailing frameworks are included for comparison:
\begin{itemize}
	\item Local learning (LL): Each client trains a deep learning network based on its pixel-level labeled data.
	\item Federated Averaging (FedAvg): All clients, owning pixel-level labels, collaboratively train a federated model.
	\item Semi-supervised federated learning via self-training~\cite{ssl_seg_fl} (FedST): FedST is proposed to utilize both pixel-level labeled and unlabeled data for federated training. FedST is selected as it does not require additional information sharing beyond the locally-updated model parameters.
	\item Our proposed Federated learning with mixed supervisions (FedMix): FedMix integrates various levels of labels.
\end{itemize}
The performance of FedAvg under the fully-supervised setting is regarded as an upper bound of federated learning techniques.
We evaluate the performance of FedMix under the semi-supervised setting by comparing FedMix with FedST. We also evaluate the performance of FedMix under various settings to show how additional weak labels improve the federated model's performance.

\myparagraph{Training details.} All the networks are trained using the Adam optimizer with an initial learning rate of 1e-3 and a batch size of 16. All methods are implemented within the PyTorch framework and trained on Nvidia GeForce Titan RTX GPUs for 300 rounds. The federated training is performed synchronously and the federated model parameters are updated every training round. We set $\epsilon=0.9$, $\lambda=10$, and $\beta=1.5$ and $\beta=3$ for adaptive aggregation on breast tumor and skin lesion segmentation respectively. 

{\renewcommand{\arraystretch}{1.0}
	\begin{table}[t]
		\centering
		\caption{Quantitative results of local learning (LL) and FedAvg under the fully-supervised setting for breast tumor segmentation.}\label{breast_full}
		\begin{tabular}{c|c|c|c|c}
			\hline
			\multirow{2}{*}{Frameworks} & C1             & C2             & C3             & \multirow{2}{*}{Avg.} \\ \cline{2-4}
			& $L$              & $L$              & $L$              &                       \\ \hline
			LL                          & 66.96          & 87.37          & 87.23          & 80.52                 \\ \hline
			FedAvg                          & \textbf{77.46} & \textbf{92.44} & \textbf{87.12} & \textbf{85.67}        \\ \hline
		\end{tabular}
	\end{table}
}

{\renewcommand{\arraystretch}{1.0}
	\begin{table}[t]
		\centering
		\caption{Quantitative results of different learning frameworks under the semi-supervised setting for breast tumor segmentation.}\label{fair_comparison_breast}
		\begin{tabular}{c|c|c|c|c}
			\hline
			\multirow{2}{*}{Frameworks}   & C1                         & C2                         & C3                         & \multirow{2}{*}{Avg.}     \\ \cline{2-4}
			& $U$                          & $U$                         & $L$                          &                           \\ \hline
			LL (trained on C3)                            & 64.72                      & 83.40                      & 87.23                      & 78.45                     \\ \hline
			FedST~\cite{ssl_seg_fl} & \multicolumn{1}{l|}{64.83} & \multicolumn{1}{l|}{85.66} & \multicolumn{1}{l|}{86.38} & \multicolumn{1}{l}{78.95} \\ \hline
			FedMix                        & \textbf{68.17}             & \textbf{89.19}             & \textbf{87.97}             & \textbf{81.77}            \\ \hline
		\end{tabular}
	\end{table}
}

\subsection{Results on Breast Tumor Segmentation}\label{breast_exp}

\myparagraph{Experiment settings.} Data from BUS, BUSIS, and UDIAT are represented by C1, C2, and C3 respectively. To better demonstrate the value of weak labels, C3, owning the least amount of data, is selected as the client with pixel-level labels. The levels of the labels on C1 and C2 are adjusted accordingly for different cases. As only C1 contains both healthy and cancerous images, it is regarded as the client with image-level labels when needed.

{\renewcommand{\arraystretch}{1.0}
	\begin{table}[t]
		\centering
		\caption{Quantitative results of FedMix under various weakly-supervised settings for breast tumor segmentation.}\label{flos_comparison_breast}
		\begin{tabular}{c|c|c|c|c}
			\hline
			supervision      & \multirow{2}{*}{C1} & \multirow{2}{*}{C2} & \multirow{2}{*}{C3} & \multirow{2}{*}{Avg.} \\ \cline{1-1}
			{[}C1, C2, C3{]} &                     &                     &                     &                       \\ \hline
			{[}$U$, $U$, $L${]}    & 68.17               & 89.19               & 87.97               & 81.77                 \\ \hline
			{[}$I$, $U$, $L${]}    & 68.37               & \textbf{89.47}      & 88.56               & 82.13                 \\ \hline
			{[}$B$, $B$, $L${]}    & \textbf{71.26}      & 89.36               & \textbf{89.41}      & \textbf{83.34}        \\ \hline
		\end{tabular}
	\end{table}
}

{\renewcommand{\arraystretch}{1.0}
	\begin{table}[t]
		\centering
		\caption{Quantitative results of federated learning under the fully-supervised setting with various aggregation functions for breast tumor segmentation. AdaptAgg is the proposed aggregation function.}\label{fl_aa_breast}
		\begin{tabular}{ccccc}
			\hline
			\multicolumn{1}{c|}{Aggregation} & \multicolumn{1}{c|}{C1}             & \multicolumn{1}{c|}{C2}             & \multicolumn{1}{c|}{C3}             & \multirow{2}{*}{Avg.} \\ \cline{2-4}
			\multicolumn{1}{c|}{Function}                                    & \multicolumn{1}{c|}{$L$}              & \multicolumn{1}{c|}{$L$}              & \multicolumn{1}{c|}{$L$}              &                       \\ \hline
			\multicolumn{1}{c|}{FedAvg}                                  & \multicolumn{1}{c|}{77.46} & \multicolumn{1}{c|}{\textbf{92.44}} & \multicolumn{1}{c|}{87.12} & 85.67        \\ \hline
			\multicolumn{1}{c|}{AdaptAgg}   & \multicolumn{1}{c|}{\textbf{79.02}} & \multicolumn{1}{c|}{93.08} & \multicolumn{1}{c|}{\textbf{88.27}} & \textbf{86.79}       \\ \hline
		\end{tabular}
	\end{table}
}

\myparagraph{Quantitative evaluation.}
According to Table~\ref{breast_full}, \eg, in the fully-supervised setting, the LL model of C1 has the lowest DC of 66.96\%, indicating the large intra-client variations among its data. C2 and C3 performs better than C1, \ie, 87.37\% and 87.23\% respectively. With FedAvg, every client benefits from the federation, especially C1 with an increase of 10.50\% in DC. Besides, training a federated model with more data from different clients is useful to learn more generalizable features, leading to an average increase of 5.15\% in DC. 

\begin{figure*}[t!]
	\centering
	\includegraphics[width=1\columnwidth]{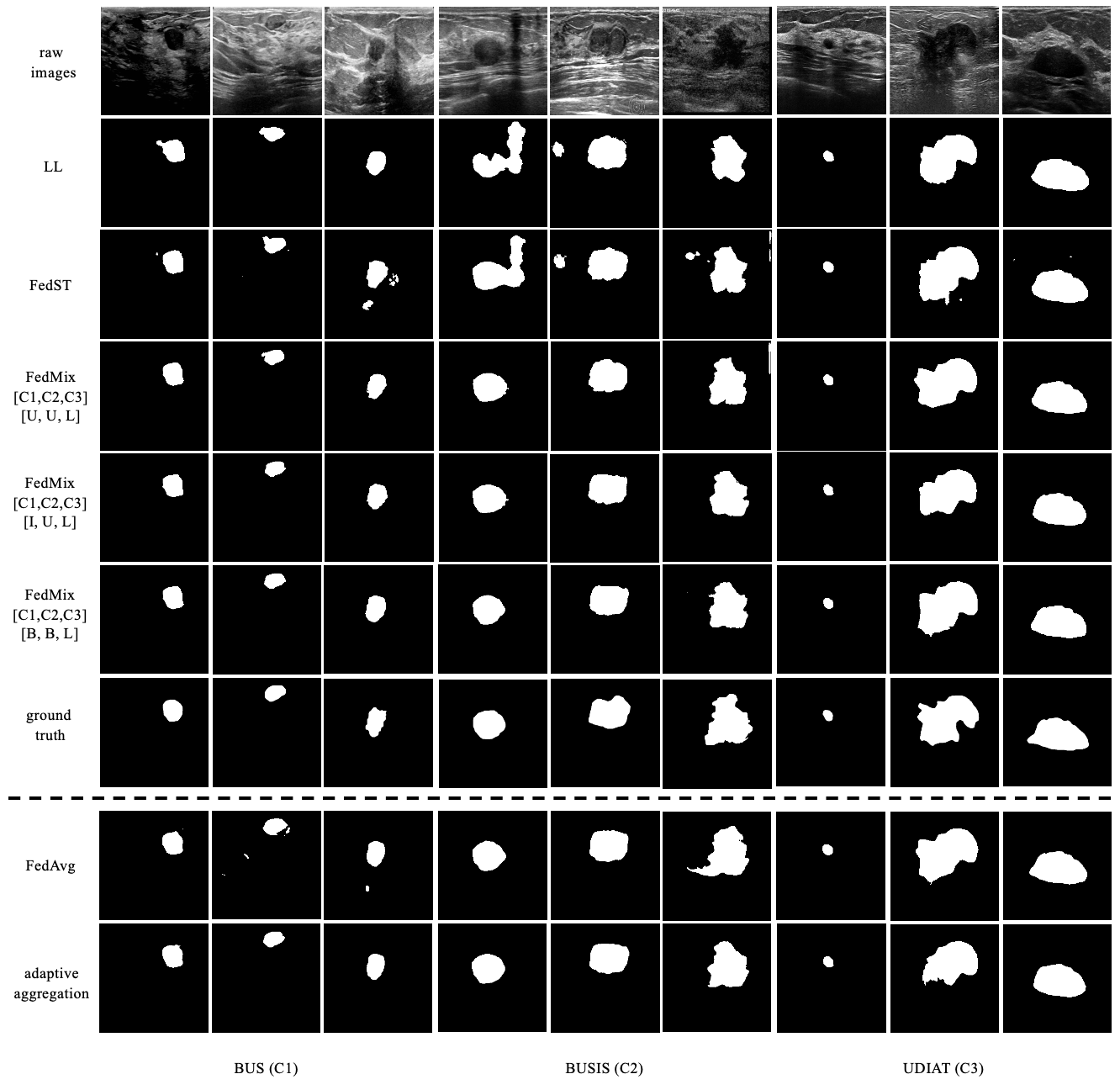}
	\caption{
		Exemplar qualitative results of different learning frameworks for breast tumor segmentation. \textbf{The upper part} (Rows 1 to 7): the raw images, the segmentation maps produced by local learning (LL), FedST and FedMix under semi-supervision (\ie,[C1, C2, C3] = [$U$, $U$, $L$]), the segmentation maps of FedMix under mixed supervision (\ie,[C1, C2, C3] = [$I$, $U$, $L$] and [C1, C2, C3] = [$B$, $B$, $L$]), and the manual annotations by experts respectively. \textbf{The lower part} (Rows 8 to 9): the segmentation maps obtained by federated learning under full pixel-level supervision using FedAvg and the proposed adaptive aggregation function respectively.
	}\label{qual_breast}
\end{figure*}

Quantitative results of FedMix and FedST under the semi-supervised setting are provided in Table.~\ref{fair_comparison_breast}. For LL, the results of C1 and C2 are produced using the model trained on C3. Compared to the locally-learned models under the fully-supervised setting in Table~\ref{breast_full}, there exists slight performance degradation on C1 and C2, \ie, 2.24\% and 3.97\% decrease in DC respectively, indicating the limitation of the model trained only on C3. By utilizing the unlabeled data on C1 and C2 for training, FedST and FedMix are able to train better federated models compared to LL. The overall improvements of FedST are quite limited with an average increase of 0.50\% in DC while the segmentation results on C3 are badly affected. Comparatively, FedMix consistently improves the results of all the three clients, leading to an average increase of 3.32\% and 2.82\% in DC for LL and FedST respectively. 

One interesting observation is that FedMix in semi-supervised learning outperforms LL with full supervisions, demonstrating the effectiveness of FedMix in exploring hidden information in unlabeled data. Quantitative results of FedMix under different settings are presented in Table~\ref{flos_comparison_breast}. When C1 owns image-level labels, not only C1 but also C2 and C3 would benefit from the federation, shown by performance improvements across clients, \ie, an average of 0.36\% increase in DC. When C1 and C2 have access to bounding box labels, the DC scores of C1 and C3 are further improved, with an average increase of 1.57\% and 1.11\% compared to FedMix with weaker supervisions. To validate the effectiveness of adaptive aggregation, we compare FedAvg and adaptive aggregation under the fully-supervised setting. The results are presented in Table~\ref{fl_aa_breast}. Putting more emphasis on more reliable clients via adaptive aggregation effectively improves the DC by 1.12\%.  

\myparagraph{Qualitative evaluation.}
According to Fig.~\ref{qual_breast}, LL on C3 produces quite a few false positives on C2, indicating poor generalization capability due to limited training data. 
Under the semi-supervised setting, though the unlabeled data of C1 and C2 is used for training, the segmentation results of FedST are close to those of LL as learning from incorrect pseudo labels is not helpful and may be detrimental. Comparatively, FedMix can utilize the useful information in unlabeled data and the model generates predictions close to the experts' annotations. The introduction of stronger supervision signals (\ie, from $U$ to $I$ and $B$) to FedMix would further reduce false positives and improve the shape preservation of tumor regions. 
The utilization of adaptive aggregation in federated learning is beneficial even under the fully-supervised setting. Adaptively aggregated federated model can better capture the boundaries and shapes of the tumor regions and contain fewer false positives compared to the model learned using FedAvg. 

\subsection{Results on Skin Lesion Segmentation}\label{skin_exp}
	
\myparagraph{Experiment setting.} 
Images from Rosendahl, Vidir-modern, Vidir-old, and Vidir-molemax are represented by C1, C2, C3, and C4 respectively, and C3, owning the least amount of data, is selected as the client with pixel-level labels. The levels of the labels on C1, C2, and C4 are adjusted accordingly for different cases. 

\myparagraph{Quantitative results.}
From Table~\ref{skin_full}, under the fully-supervised setting, FedAvg improves the performance of the locally-learned models by an average of 0.96\% in DC, indicating that cross-client collaboration is beneficial. 

{\renewcommand{\arraystretch}{1.0}  
	\begin{table}[t]
		\centering
		\caption{Quantitative results of local learning (LL) and FedAvg under the fully-supervised setting for skin lesion segmentation.}\label{skin_full}
		\begin{tabular}{c|c|c|c|c|c}
			\hline
			\multirow{2}{*}{Frameworks} & C1             & C2             & C3             & C4             & \multirow{2}{*}{Avg.} \\ \cline{2-5}
			& $L$              & $L$              & $L$              & $L$              &                       \\ \hline
			LL                          & 88.98          & 93.21          & 94.33          & 94.93          & 92.86                 \\ \hline
			FedAvg                          & \textbf{90.39} & \textbf{93.57} & \textbf{95.88} & \textbf{95.44} & \textbf{93.82}        \\ \hline
		\end{tabular}
	\end{table}
}

{\renewcommand{\arraystretch}{1.0}
	\begin{table}[t]
		\centering
		\caption{Quantitative results of different learning frameworks under the semi-supervised setting for skin lesion segmentation.}\label{skin_unlabeled}
		\begin{tabular}{c|c|c|c|c|c}
			\hline
			\multirow{2}{*}{Frameworks} & C1             & C2             & C3             & C4             & \multirow{2}{*}{Avg.} \\ \cline{2-5}
			& $U$              & $U$              & $L$              & $U$              &                       \\ \hline
			LL (trained on C3)                           & 74.55          & 72.85          & 94.33          & 91.21          & 83.23                 \\ \hline
			FedST~\cite{ssl_seg_fl}                    & 75.08          & 74.08          & 93.78          & \textbf{92.24} & 83.79                 \\ \hline
			FedMix                      & \textbf{80.55} & \textbf{81.72} & \textbf{94.54} & 90.92          & \textbf{86.93}        \\ \hline
		\end{tabular}
	\end{table}
}

The key for semi-supervised federated learning is to extract and use accurate information from the unlabeled data. Under the semi-supervised setting, where only C3 has access to annotation (\ie, $L$), we present the results in Table~\ref{skin_unlabeled}. The locally-learned (LL) model on C3 does not perform well on C1 and C2, observed through the significant performance degradation which indicates severe inter-client variations between \{C3, C4\} and \{C1, C2\}. As a result, the pseudo labels on \{C1, C2\} generated by the model trained on C3 may be inaccurate, utilizing which for training would be harmful. Instead of using all the pseudo labels, FedST makes use of only confident predictions. While the model learned through FedST has an average of 0.56\% increase in DC compared to LL, it performs worse on C3, \ie, 0.55\% decrease in DC. The performance drop may disincentive C3 to participate in the federation thus hindering the deployment of FedST. With dynamic sample selection and adaptive aggregation, FedMix manages to select high-quality unlabeled data and more accurate pseudo labels for training, thus improving the segmentation performance on C3. Additionally, compared to LL, both C1 and C2 obtain significant performance improvements with an average increase of 6.00\% and 8.87\% in DC respectively. In general, FedMix achieves better overall performance, resulting in an average increase of 3.14\% in DC compared to FedST.

{\renewcommand{\arraystretch}{1.0}
\begin{table}[t]
		\centering
		\vspace{2mm}
		\caption{Quantitative results of FedMix under various mixed supervised settings for skin lesion segmentation.}\label{skin_unified}
		\begin{tabular}{c|c|c|c|c|c}
			\hline
			Supervision          & \multirow{2}{*}{C1} & \multirow{2}{*}{C2} & \multirow{2}{*}{C3} & \multirow{2}{*}{C4} & \multirow{2}{*}{Avg.} \\ \cline{1-1}
			{[}C1, C2, C3, C4{]} &                     &                     &                     &                     &                       \\ \hline
			{[}$U$, $U$, $L$, $U${]}     & 80.55               & 81.72               & 94.54               & 90.92               & 86.93                 \\ \hline
			{[}$B$, $B$, $L$, $B${]}     & \textbf{88.80}       & \textbf{93.11}      & \textbf{95.82}      & \textbf{94.41}      & \textbf{93.04}        \\ \hline
		\end{tabular}
	\end{table}
}

{\renewcommand{\arraystretch}{1.0}
	\begin{table}[t]
		\centering
		\caption{Quantitative results under the fully-supervised setting with various aggregation functions for skin lesion segmentation. AdaptAgg is the proposed adaptive aggregation.}\label{fl_aa_skin}
		\begin{tabular}{cccccc}
			\hline
			\multicolumn{1}{c|}{Aggregation} & \multicolumn{1}{c|}{C1}             & \multicolumn{1}{c|}{C2}             & \multicolumn{1}{c|}{C3}             & \multicolumn{1}{c|}{C4}             & \multirow{2}{*}{Avg.} \\ \cline{2-5}
			\multicolumn{1}{c|}{Function}                                    & \multicolumn{1}{c|}{$L$}              & \multicolumn{1}{c|}{$L$}              & \multicolumn{1}{c|}{$L$}              & \multicolumn{1}{c|}{$L$}              &                       \\ \hline
			\multicolumn{1}{c|}{FedAvg}                                  & \multicolumn{1}{c|}{90.39} & \multicolumn{1}{c|}{93.57} & \multicolumn{1}{c|}{95.88} & \multicolumn{1}{c|}{95.44}            & 
			93.82       \\ \hline
			\multicolumn{1}{c|}{AdaptAgg}                                   & \multicolumn{1}{c|}{\textbf{90.91}} & \multicolumn{1}{c|}{\textbf{93.73}} & \multicolumn{1}{c|}{\textbf{96.78}} & \multicolumn{1}{c|}{\textbf{95.51}} & \textbf{94.23}      \\ \hline
		\end{tabular}
		\vspace{-3mm}
	\end{table}
}

\begin{figure*}[t!]
	\centering
	\includegraphics[width=1\columnwidth]{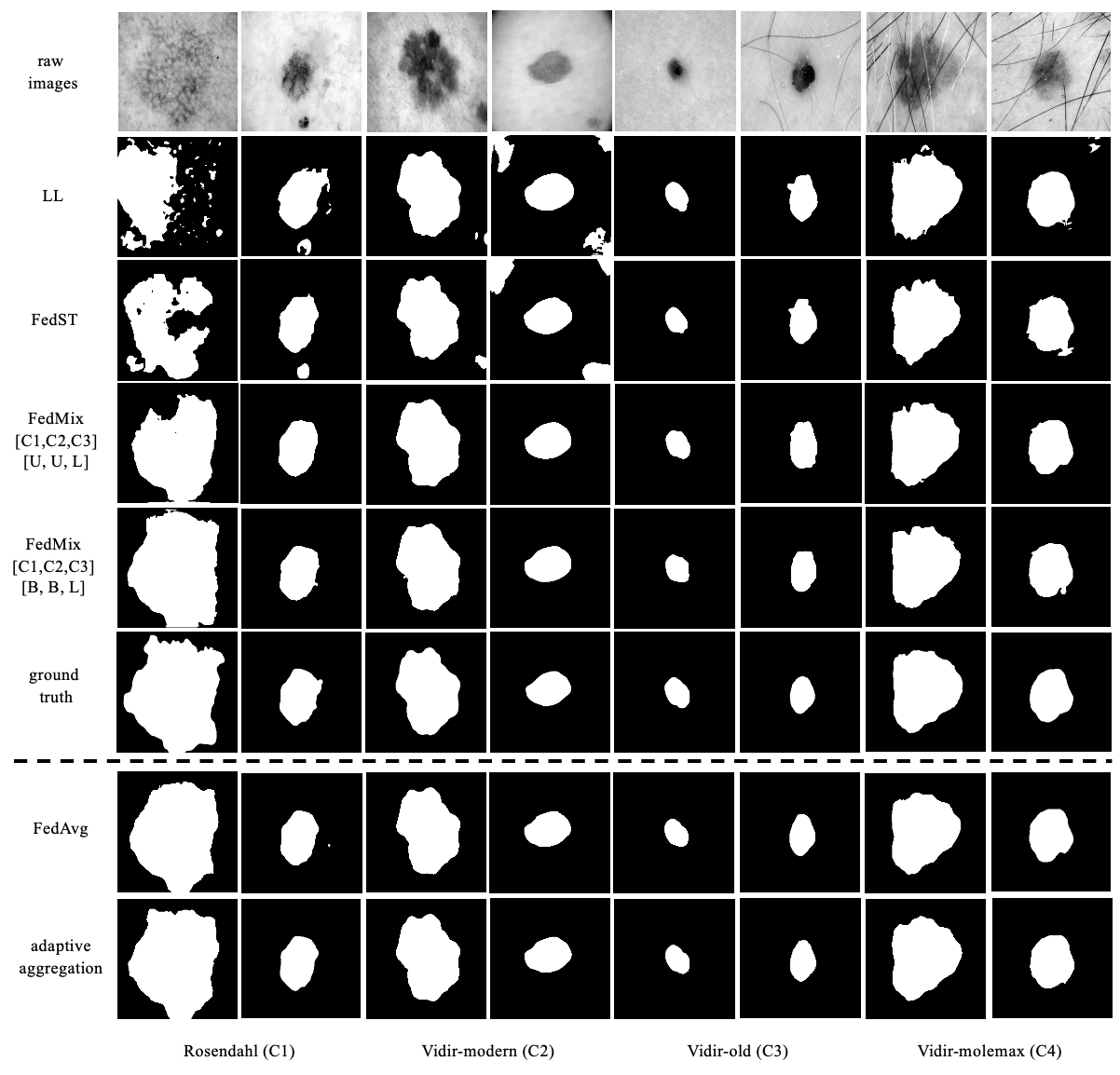}
	\caption{Qualitative results of different learning frameworks for skin lesion segmentation. \textbf{The upper part} (Rows 1 to 6): the raw images, the segmentation maps produced by local learning (LL), FedST, FedMix under semi-supervision (\ie,[C1, C2, C3, C4] = [$U$, $U$, $L$, $U$]), FedMix under mixed supervision (\ie,[C1, C2, C3, C4] = [$B$, $B$, $L$, $B$]), and the expert annotations respectively. \textbf{The lower part }(Rows 7 to 8): the segmentation maps obtained by federated learning under the fully-supervised setting with FedAvg and the proposed adaptive aggregation function respectively.
	}\label{qual_skin}
\end{figure*}

Quantitative results of FedMix under various settings are presented in Table~\ref{skin_unified}. Incorporating bounding box labels for training improves the pseudo labels' accuracy. Consequently, the segmentation performance of FedMix is further improved by 6.11\%, approaching the performance of FedAvg under the fully-supervised setting. Bounding box labels are much easier to obtain than pixel-level labels, making FedMix more valuable in clinical scenarios. We further conduct a comparison between FedAvg and adaptive aggregation under the fully-supervised setting, presented in Table~\ref{fl_aa_skin}. The proposed adaptive aggregation function can better utilize the high-quality data and balance the weights among clients, leading to better convergence and segmentation performance. 

\myparagraph{Qualitative results.}
Qualitative results of skin lesion segmentation are shown in Fig.~\ref{qual_skin}. Consistent with the quantitative results, the segmentation maps on C1 and C2, produced by the locally-learned model on C3, are inaccurate, due to large inter-client variations between \{C1, C2\} and \{C3, C4\}. While the segmentation maps produced by FedST are slightly more accurate compared to LL, learning from confident pseudo labels is insufficient to train a generalizable model, shown through the inaccurate segmentation maps produced by FedST on C1 and C2. Under the same supervision setting, FedMix produces more accurate segmentation maps by dynamically selecting the high-quality pseudo labels for training. Given stronger supervisions, \eg, bounding box labels, FedMix improves the segmentation quality, especially on tumor shape preservation. Through the comparison under the fully-supervised setting, we observe that the segmentation maps produced by adaptive aggregation contain fewer false negatives and have better shape consistencies with manual annotations compared to FedAvg.
\section{Ablation Studies}\label{discussion}
\subsection{Effectiveness of Dynamic Sample Selection} \label{consistencies_reliable} 
We remove the label refinement step in FedMix and utilize FedAvg for comparison. Quantitative results are presented in Table~\ref{abl_freeze}. We can observe that without dynamic sample selection, the model may learn from incorrect pseudo labels which is detrimental for convergence. Dynamic sample selection is based on the intuition where the prediction consistencies between the two models given the same input image are positively correlated with the accuracy of the pseudo labels. We perform separate evaluations on the three datasets for breast tumor segmentation, (\ie, BUS (C1), BUSIS (C2), and UDIAT (C3)). For each client, we train two differently initialized models, $F_{1}$ and $F_{2}$, locally on 80\% of the data for 20 training rounds. 

The prediction consistencies between the two models, measured in DC (\%), are used to select the evaluation set from the remaining 20\% of the data according to the consistency threshold $\epsilon$. With a smaller $\epsilon$, more samples with lower prediction consistencies are included for evaluation. With the increase of $\epsilon$, as only the samples with high prediction consistencies are selected, the overall DC accuracy is higher. The findings in Table~\ref{ablation_eps} validate our assumption and demonstrate the value of dynamic sample selection in filtering inaccurate pseudo labels during training.  

{\renewcommand{\arraystretch}{1.0}
	\begin{table}[t]
		\centering
		\caption{Quantitative results of FedMix with and without dynamic sample selection for breast tumor and skin lesion segmentation.}\label{abl_freeze}
		\begin{tabular}{cccccc}
			\hline
			\multicolumn{1}{c|}{Sample} & \multicolumn{1}{c|}{C1}             & \multicolumn{1}{c|}{C2}             & \multicolumn{1}{c|}{C3}             & \multicolumn{1}{c|}{C4}             & \multirow{2}{*}{Avg.} \\ \cline{2-5}
			\multicolumn{1}{c|}{Selection}                                    & \multicolumn{1}{c|}{$U$}              & \multicolumn{1}{c|}{$U$}              & \multicolumn{1}{c|}{$L$}              & \multicolumn{1}{c|}{$U$}              &                       \\ \hline
			\multicolumn{6}{c}{Breast tumor segmentation}                                                                                                                                                                                            \\ \hline
			\multicolumn{1}{c|}{$\times$}                                   & \multicolumn{1}{c|}{34.92}          & \multicolumn{1}{c|}{47.69}          & \multicolumn{1}{c|}{30.41}          & \multicolumn{1}{c|}{N/A}            & 37.67                 \\ \hline
			\multicolumn{1}{c|}{\checkmark}                                   & \multicolumn{1}{c|}{\textbf{66.92}} & \multicolumn{1}{c|}{\textbf{88.49}} & \multicolumn{1}{c|}{\textbf{86.95}} & \multicolumn{1}{c|}{N/A}            & \textbf{80.78}        \\ \hline
			\multicolumn{6}{c}{Skin lesion segmentation}                                                                                                                                                                                             \\ \hline
			\multicolumn{1}{c|}{$\times$}                                   & \multicolumn{1}{c|}{45.38}          & \multicolumn{1}{c|}{33.10}          & \multicolumn{1}{c|}{55.11}          & \multicolumn{1}{c|}{41.28}          & 43.27                 \\ \hline
			\multicolumn{1}{c|}{\checkmark}                                   & \multicolumn{1}{c|}{\textbf{81.30}} & \multicolumn{1}{c|}{\textbf{78.10}} & \multicolumn{1}{c|}{\textbf{94.43}} & \multicolumn{1}{c|}{\textbf{91.11}} & \textbf{86.24}        \\ \hline
		\end{tabular}
	\end{table}
}
	
{\renewcommand{\arraystretch}{1.0}
	\begin{table}[t]
		\centering
		\caption{The effect of the threshold $\epsilon$ to the quantitative results (DC \%) on each client for breast tumor segmentation.}
		\label{ablation_eps}
		\begin{tabular}{c|c|c|c}
			\hline
			$\epsilon$ & BUS (C1) & BUSIS (C2) & UDIAT (C3) \\ \hline
			0.1 & 12.6 & 13.5 & 18.8 \\ \hline
			0.2 & 22.0 & 22.9 & 21.3 \\ \hline
			0.3 & 25.3 & 39.8 & 26.5 \\ \hline
			0.4 & 56.1 & 40.2 & 45.5 \\ \hline
			0.5 & 55.9 & 45.2 & 40.7 \\ \hline
			0.6 & 66.1 & 60.9 & 52.9\\ \hline
			0.7 & 66.2 & 73.6 & 64.5\\ \hline
			0.8 & 72.3 & 77.0 & 64.9\\ \hline
			0.9 & \textbf{86.07} & \textbf{89.1} & \textbf{79.6} \\ \hline
		\end{tabular}
	\end{table}
}
{\renewcommand{\arraystretch}{1.0}
	\begin{table}[t!]
		\centering
		\caption{Quantitative results of FedMix with and without adaptive aggregation for breast tumor and skin lesion segmentation.}\label{abl_aa}
		\begin{tabular}{cccccc}
			\hline
			\multicolumn{1}{c|}{Adaptive} & \multicolumn{1}{c|}{C1}             & \multicolumn{1}{c|}{C2}             & \multicolumn{1}{c|}{C3}             & \multicolumn{1}{c|}{C4}             & \multirow{2}{*}{Avg.} \\ \cline{2-5}
			\multicolumn{1}{c|}{Aggregation}                                     & \multicolumn{1}{c|}{$U$}              & \multicolumn{1}{c|}{$U$}              & \multicolumn{1}{c|}{$L$}              & \multicolumn{1}{c|}{$U$}              &                       \\ \hline
			\multicolumn{6}{c}{Breast tumor segmentation}                                                                                                                                                                                             \\ \hline
			\multicolumn{1}{c|}{$\times$}                                    & \multicolumn{1}{c|}{66.92}          & \multicolumn{1}{c|}{88.49}          & \multicolumn{1}{c|}{86.95}          & \multicolumn{1}{c|}{N/A}            & 80.78                 \\ \hline
			\multicolumn{1}{c|}{\checkmark}                                    & \multicolumn{1}{c|}{\textbf{68.17}} & \multicolumn{1}{c|}{\textbf{89.19}} & \multicolumn{1}{c|}{\textbf{87.97}} & \multicolumn{1}{c|}{N/A}            & \textbf{81.78}        \\ \hline
			\multicolumn{6}{c}{Skin lesion segmentation}                                                                                                                                                                                              \\ \hline
			\multicolumn{1}{c|}{$\times$}                                    & \multicolumn{1}{c|}{\textbf{81.30}} & \multicolumn{1}{c|}{78.10}          & \multicolumn{1}{c|}{94.43}          & \multicolumn{1}{c|}{\textbf{91.11}} & 86.24                 \\ \hline
			\multicolumn{1}{c|}{\checkmark}                                    & \multicolumn{1}{c|}{80.55}          & \multicolumn{1}{c|}{\textbf{81.72}} & \multicolumn{1}{c|}{\textbf{94.54}} & \multicolumn{1}{c|}{90.92}          & \textbf{86.93}        \\ \hline
		\end{tabular}
	\end{table}
}

\subsection{Effectiveness of Adaptive Aggregation}\label{value_aa}
We compare adaptive aggregation with FedAvg and present the results in Table~\ref{abl_aa}. For breast tumor segmentation, adaptive aggregation consistently improves performance across clients, with an average of 1.00\% increase in DC compared to FedAvg. For skin lesion segmentation, due to the inter-client variations between \{C1, C2\} and \{C3, C4\}, adaptive aggregation focuses more on minimizing the training losses on C1 and C2. As a result, the average DC increase of \{C1, C2\} is 1.44\% while the corresponding increase on C4 is limited to 0.19\%. Overall, adaptive aggregation outperforms FedAvg. 
Till now, aggregation weight optimization in federated learning is still an open problem and should be further explored in the future.

\section{Conclusion}\label{conclusion}
FedMix is the first federated learning framework that makes effective use of different levels of labels on each client for medical image segmentation. In FedMix, we first generate pseudo labels from clients and use supervision-specific refinement strategies to improve the accuracy and quality of pseudo labels. Then the high-quality data of each client is selected through dynamic sample selection for local model updates. To better update the federated model, FedMix utilizes an adaptive aggregation function to adjust the weights of clients according to both data quantity and data quality. Experimental results on two segmentation tasks demonstrate the effectiveness of FedMix on learning from various supervisions, which is valuable to reduce the annotation burden of medical experts. In the semi-supervised federated setting, FedMix outperforms the state-of-the-art approach FedST. Compared to FedAvg, the proposed adaptive aggregation function achieves consistent performance improvements on the two tasks under the fully-supervised setting. We believe the methods proposed in FedMix are widely-applicable in FL for medical image analysis beyond mixed supervisions.

\end{document}